\documentclass[letterpaper, 10 pt, conference]{ieeeconf}  

\IEEEoverridecommandlockouts                              

\usepackage{graphics} 
\usepackage{epsfig} 
\usepackage{mathptmx} 
\usepackage{times} 
\usepackage{amsmath} 
\usepackage{amssymb}  
\usepackage{color}
\usepackage{graphicx}
\usepackage{url}
\usepackage{hyperref}
\usepackage{cite}
\usepackage{makecell}
\usepackage{algorithm}
\usepackage{algpseudocode}
\usepackage{bm}
\usepackage[subtle,tracking=normal]{savetrees}

\title{\LARGE \bf
Safe Bimanual Teleoperation with Language-Guided Collision Avoidance
}

\author{Dionis Totsila$^{\star}$, Clemente Donoso Krauss$^{\star}$, Enrico Mingo Hoffman, Jean-Baptiste Mouret, Serena Ivaldi
\thanks{Corresponding author: serena.ivaldi@inria.fr}%
\thanks{All the authors are with INRIA, Université de Lorraine, CNRS, 54000 Nancy, France. }
\thanks{($^{\star})$ Equal Contribution.}%
}

\begin{document}
\maketitle
\thispagestyle{empty}
\pagestyle{empty}
\begin{abstract}
Teleoperating precise bimanual manipulations in cluttered environments is challenging for operators, who often struggle with limited spatial perception and difficulty estimating distances between target objects, the robot's body, obstacles, and the surrounding environment. To address these challenges, local robot perception and control should assist the operator during teleoperation. In this work, we introduce a safe teleoperation system that enhances operator control by preventing collisions in cluttered environments through the combination of immersive VR control and voice-activated collision avoidance. Using HTC Vive controllers, operators directly control a bimanual mobile manipulator, while spoken commands such as ``\textit{avoid the yellow tool}'' trigger visual grounding and segmentation to build 3D obstacle meshes. These meshes are integrated into a whole-body controller to actively prevent collisions during teleoperation. Experiments in static, cluttered scenes demonstrate that our system significantly improves operational safety without compromising task efficiency.
\end{abstract}
\section{INTRODUCTION}

Teleoperation is a cornerstone of robotics, enabling humans to interact with hazardous or remote environments. In addition, it plays a pivotal role in the current era of data-driven robotics, where large-scale data collection is crucial for behavior cloning and learning-based control~\cite{darvish2023teleop}.

Despite these advantages, teleoperation remains challenging in complex, cluttered environments. While immersive VR setups now enable high-fidelity, direct control of bimanual robots~\cite{cheng2024opentelevisionteleoperationimmersiveactive,penco2023prescient}, humans are still prone to error. A moment's inattention or slight misjudgment, stemming from limited spatial awareness or poor distance estimation, can result in unintended collisions, risking damage to both the environment and the robot.

Traditionally, teleoperation systems have relied primarily on visual and haptic feedback to avoid collisions, placing significant cognitive demands on the operator. We argue that allowing operators to express intent through natural language, by specifying which objects to avoid, can make teleoperation far more intuitive. With recent advances in visual and language models, it has become feasible to dynamically "program" the robot controller via verbal commands to enforce task-relevant collision avoidance.

\begin{figure}[t!]
    \centering
    \includegraphics[width=\linewidth]{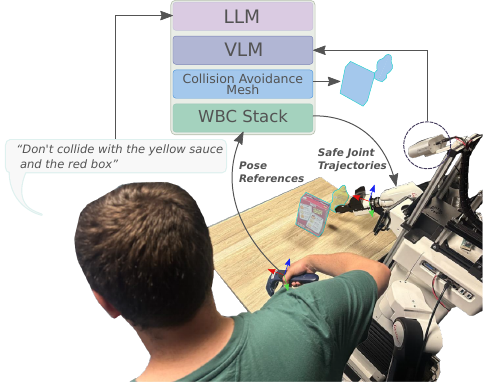}
    \vspace{-1.5em}
    \caption{The operator teleoperates both arms of the robot using VR controllers while instructing the system via speech to avoid the ``\textit{yellow sauce}'' and the ``\textit{red box}''. This enables intent-aware collision avoidance.}
    \label{fig:concept}
\end{figure}

Conventional collision-avoidance methods typically depend on either a precise, continuously updated model of the environment with manually labeled obstacles, or they conservatively treat all 3D sensor observations as potential hazards. These approaches are impractical for dexterous manipulation tasks, where scenes are highly dynamic, object configurations change frequently, and the relevance of obstacles is inherently task- and context-dependent. Moreover, what constitutes a collision risk is fluid, an object that should be avoided at one moment may become the target of manipulation the next. Recent work highlights the limitations of static or purely distance-based risk models, advocating instead for approaches that incorporate motion dynamics, acceleration, and task context to flexibly assess collision risk during teleoperation~\cite{gautam2025intelligentcollisionavoidance, zhang2024collisionriskassessment}.

Simultaneously, advances in large language models (LLMs), vision-language models (VLMs), and vision-language-action (VLA) systems are bringing natural language interaction to teleoperation~\cite{totsila2024words2contact, kim2024openvla, zhang2023llmhrireview}. However, for many real-world manipulation tasks, especially in bimanual teleoperation, full task specifications cannot be easily articulated in advance. Instead, key decisions emerge in real-time as operators interact with the environment. In such scenarios, language serves as an intuitive, hands-free channel for operators to dynamically communicate intent and constraints (e.g., ``\textit{avoid the yellow tool}'') without interrupting control.

Building on these insights, we introduce a general-purpose bimanual teleoperation framework that integrates immersive VR-based 6-DoF control with speech-driven, intent-aware collision avoidance. Operators control both arms of a mobile manipulator via VR controllers, while issuing natural language commands such as ``\textit{avoid the gray box}'' to trigger visual grounding and real-time collision avoidance (Fig.~\ref{fig:concept}).
\begin{figure*}[th!]
    \centering
    \includegraphics[width=0.8\textwidth]{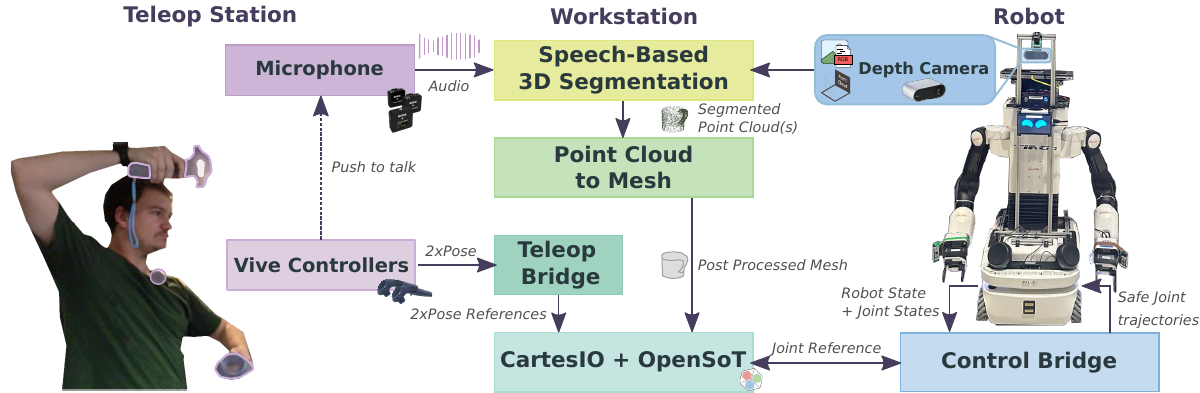}
    \caption{Illustration of the main components of the system. The teleoperation station, equipped with a microphone, captures the operator's verbal input, which is processed by the speech-based 3D segmentation module. All computationally intensive operations are performed on a desktop with two consumer-grade GPUs. The generated collision-avoidance meshes and joint reference commands are transmitted to the control bridge on the robot. The robot is equipped with an RGB-D camera mounted on its head.}
    \label{fig:pipeline_overview}
\end{figure*}

We evaluate our system through a pilot study in which five participants performed a standardized pick-and-place task in a narrow, cluttered environment. For each trial, we compare the outcomes of user-generated motions executed with and without speech-augmented collision avoidance. Control commands recorded during live trials were replayed with collision avoidance disabled to assess its impact. While no collisions occurred during live operation, four out of five replayed trials resulted in collisions, highlighting the system's effectiveness in ensuring safe teleoperation. Beyond the pilot study, we demonstrate the system's flexibility and generalizability across a variety of challenging downstream tasks, which are showcased in Fig.~\ref{fig:downstream_tasks} and in the accompanying video.


\section{SYSTEM OVERVIEW}
Our system and its components are illustrated in Fig.~\ref{fig:pipeline_overview}. The architecture is divided between the teleoperation station, which includes a workstation equipped with two consumer-grade GPUs, and the onboard computer of the robot. Each key component is described in detail in the following sections.

\subsection{LLM \& VLM-based 3D Object Segmentation}
\label{sec:vlm_seg}
Our system uses a speech-driven segmentation pipeline to identify the sources of collisions. It leverages vision-language models (VLMs) to ground verbal commands in the scene and extract corresponding 3D object regions in real time.

\subsubsection{Speech-to-Text Object Phrase Extraction}

The segmentation pipeline is triggered when the teleoperator holds down a designated button while speaking. Audio is buffered during the press and passed to the Whisper~\cite{radford2023whisper} speech recognition model upon release. The resulting transcript is then sent to a Large Language Model (LLaMA 3, 8B), which parses the command and returns a structured JSON indicating which objects should have collision avoidance enabled or disabled.

An LLM is essential here because teleoperation commands are often ambiguous, vary in phrasing, and may evolve over time. For instance, a user might say ``\textit{add collision for the red sauce}'' then later ``\textit{remove ketchup}'', referring to the same object. Handling such variability and intent shifts with rigid rules is fragile. The LLM also helps recover from transcription errors like ``\textit{mustard}'' being misheard as ``\textit{monster}'', using contextual reasoning and knowledge of the scene to infer likely intent.

This approach allows natural, flexible voice interaction while producing structured outputs that can be directly used for dynamic collision avoidance control.

\subsubsection{Open-Vocabulary Object Detection}
We use Grounding DINO~\cite{liu2023grounding} a vision-language model trained for open-vocabulary phrase grounding, to detect the referenced objects in the current RGB image of the scene. Unlike traditional object detectors constrained to fixed label sets, Grounding DINO accepts free-form text and returns 2D bounding boxes, enabling robust handling of generic descriptions such as  ``\textit{the green round thing}''.

This makes the system well-suited for open-world and unstructured environments, but also more accessible to any user, as different people might refer to the same thing in different ways.

Each bounding box is then passed to the Segment Anything Module (SAM)~\cite{kirillov2023segment}, which generates high-resolution 2D segmentation masks. Although open-vocabulaty segmentation models like CLIPSeg~\cite{luddecke2022image} exist, we found that this two-step detection-segmentation approach yields superior results in terms of mask quality.

Since our system directly uses a real-time registered point cloud from the robot's RGB-D camera\footnote{We use the Orbbec Femto Bolt Camera.}, we can project the 2D masks into 3D space using the known RGB image and point cloud correspondence. These sub-clouds are then passed to a post-processing pipeline (see Sec.~\ref{sec:pc_processing}).
\label{sec:pc_processing}

\subsubsection{Point Cloud Processing}
\begin{figure}[bhtp!]
    \centering
    \includegraphics[width=\linewidth]{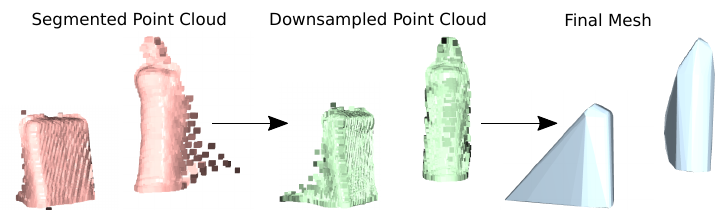}
    \vspace{-1.5em}
    \caption{From segmented point cloud to collision avoidance mesh. The segmented point cloud (left) is downsampled (center) and converted into a mesh (right) used by the robot controller for safe teleoperation.}
    \label{fig:pc_processing}
\end{figure}

Each segmented point cloud from the point cloud array is converted into an Open3D point‐cloud object~\cite{zhou2018open3dmodernlibrary3d}. Our goal is to produce a reliable collision mesh quickly, even though strict real-time reconstruction is not required. As illustrated in Fig.~\ref{fig:pc_processing}, to suppress noise and background clutter, we first apply voxel downsampling with leaf size \(\Delta_1 = 0.02\,\mathrm{m}\),
which normalizes point density and reduces computational load. Normals are then estimated via a KD‐tree neighborhood search; the search radius \(r_n\) is chosen based on density to ensure stability without mixing information from adjacent surfaces (denser clouds use \(r_n=0.10\,\mathrm{m}\), sparser ones use up to \(r_n=0.20\,\mathrm{m}\)).

Next, we employ DBSCAN clustering (\(\varepsilon=0.10\,\mathrm{m}\), \(minPts=50\)) to extract the largest spatially coherent component from the downsampled cloud, discarding isolated points and small clusters. A second voxel‐downsampling pass with leaf size \(\Delta_2 = 0.01\,\mathrm{m}\) further reduces point count while preserving essential geometric features.
Let \(N\) denote the final point count, for surface reconstruction, we use Poisson’s method~\cite{kazhdan2006poisson}, which solves for a scalar indicator function \(\chi\) whose gradient best fits the normal field:$ \nabla \cdot \nabla \chi \;=\; \nabla \cdot \mathbf{n}_i \quad\text{on the filtered point set}.$
Extracting the zero-level set of \(\chi\) yields a watertight mesh \(\mathcal{M}\). In dense cases (\(N>1.1\times10^5\)), we set octree depth \(d=12\), use a small search radius \(r=0.10\,\mathrm{m}\) to avoid blending normals across nearby surface features, and prune vertices below the 10\% density quantile. In sparser cases (\(N\le1.1\times10^5\)),
we increase \(r\) to \(0.20\,\mathrm{m}\) so that each normal estimation gathers sufficient neighbors for robustness, and prune below the 85\% quantile. We then simplify \(\mathcal{M}\) to at most 5000 triangles using quadric decimation and recompute vertex normals.

Initially, we attempted to use \(\mathcal{M}\) directly as the collision object; however, residual noise and small concavities often produced poor‐quality meshes. To mitigate this, we compute the convex hull \(\mathrm{Conv}(\mathcal{M})\) and use it as the final obstacle geometry, ensuring a clean, conservative approximation of the segmented object.

The \(\mathrm{Conv}(\mathcal{M})\) is then uniformly scaled\footnote{The scale used was 1.05.} around its centroid to provide a safety margin. Finally is transformed from the camera frame into the global \texttt{world} frame, then converted into a ROS \texttt{shape\_msgs/Mesh}, and published as a \texttt{moveit\_msgs/CollisionObject} to our whole‐body controller.

All hyperparameters voxel sizes \(\Delta_1, \Delta_2\), clustering thresholds \(\varepsilon\), \(minPts\), Poisson depth/radius \(d, r\), density thresholds, simplification target, and scale \(s\) were tuned empirically to achieve fast reconstruction times while maintaining reliable, high‐fidelity obstacle representation.

\subsection{VR Control Interface}
\label{sec:vr_control}

\begin{figure}[bthp!]
    \centering
    \includegraphics[width=0.7\linewidth]{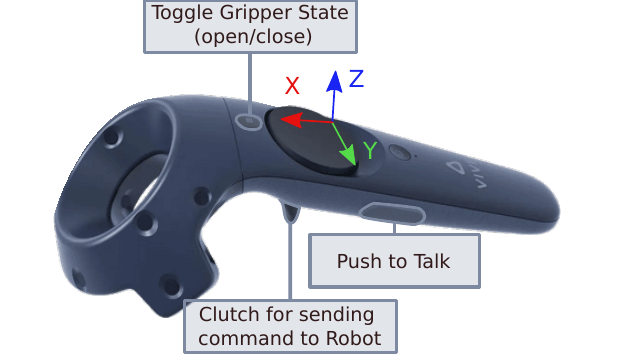}
    \caption{Custom button mappings on the HTC Vive VR controller. The controller's local reference frame is also visualized for clarity.}
    \label{fig:control_mapping}
\end{figure}

We implemented a monitor-based teleoperation system using HTC Vive controllers and OpenVR~\cite{george2023openvrteleoperationmanipulation}. Rather than wearing a Head Mounted Device (HMD), the operator views the scene on a screen, preserving awareness of the physical environment and enabling easy access to multiple camera feeds. Incidentally, this solution is more ergonomic and easier to adopt, as many people do not tolerate well the use of head mounted displays~\cite{Munafo2017}. Two handheld controllers deliver full 6-DoF Cartesian commands to the robot's end-effectors, combining intuitiveness and low cognitive load with robustness to kinematic singularities.

Tracking is provided by three Lighthouse base stations (one front, two lateral) for uninterrupted 360° coverage. Workspace calibration is performed once via a simple GUI: the operator sweeps the controllers through the workspace, the system builds a convex hull of the poses, and fits an axis-aligned bounding box. This box both constrains valid tracking and triggers haptic feedback, through a vibration on the hand-held devices, as controllers approach its limits.

All core teleoperation actions: Cartesian motion, gripper toggle, clutch, and voice-activated collision avoidance are mapped onto controller axes and buttons (see Fig.~\ref{fig:control_mapping}).

\subsection{Motion Control}
\label{subsec:cartesio}

Our system controls all upper-body DoFs of the TIAGo++ robot using an optimization-based whole-body controller (WBC), similarly to~\cite{amadio2024vocalinstructionshouseholdtasks}. The WBC receives Cartesian reference velocities from the VR teleoperation interface (Sec.~\ref{sec:vr_control}) and computes joint-level commands while respecting joint limits, and actively avoiding both self-collisions and collisions with the environment.

The controller is implemented usfing the CartesI/O framework~\cite{cartesio:icra19} and the OpenSoT library~\cite{opensot:ram24}, which together allow the control problem to be formulated and solved as a Quadratic Program (QP). Tasks represent control objectives (e.g., Cartesian tracking), while Constraints model physical limits (e.g., joint bounds or collision avoidance). These components are structured into a task stack, where higher-priority tasks take precedence over lower-priority ones.

In our setup, two velocity-level Cartesian tasks are defined for the robot's left and right end-effectors, allowing 6-DoF control of each arm. These are arranged in a soft-priority stack to allow joint motion to be shared between them when needed. A postural task in the null space of the stack regularizes redundant motions by steering the robot toward a nominal configuration.

Self-collision avoidance is implemented as a set of velocity-damping inequality constraints dynamically inserted into the QP at every control step. Following the formulation in~\cite{fang2015selfcollision}, each robot link is approximated by a capsule (sphere-swept line), and the minimum distance between relevant link pairs is computed using an extension the flexible collision library Coal~\cite{coalweb}. When this distance falls below a threshold $d_s$, a constraint is added to limit the relative velocity of the links in the direction connecting their closest points. The constraint for a link pair is defined as:
\[
\mathbf{n}_{\text{cp12}}^\top \left[ \bm{J}_{\text{cp1}} - \bm{J}_{\text{cp2}} \right] \mathbf{\dot{q}} \leq \epsilon \frac{d - d_s}{\Delta t},
\]
where $\mathbf{n}_{\text{cp12}}$ is the unit vector between closest points $\text{cp1}$ and $\text{cp2}$ on the capsule surfaces, $d$ is their distance, $\Delta t$ is the control period, and $\epsilon$ is a gain to modulate damping strength. The vector $\mathbf{\dot{q}}$ represents the joint velocity vector. The Jacobians $\bm{J}_{\text{cp1}}, \bm{J}_{\text{cp2}}$ are computed from the link Jacobians via fixed transformations. These constraints ensure the links slow down smoothly as they approach each other, preventing collision.

For object-level collision avoidance, we adopt a similar velocity-damping strategy, but instead of modeling external obstacles as part of the robot's own geometric model, we directly use the meshes generated by the speech-driven 3D segmentation pipeline (see Sec.~\ref{sec:vlm_seg}). As the operator specifies new obstacles via voice commands, the system continuously monitors the distance between the robot's links and these scene-derived meshes. Whenever a link nears one of these segmented objects within the safety threshold, a corresponding velocity-damping constraint is applied between the link and the obstacle mesh. This approach enables responsive, intent-aware obstacle avoidance and allows the robot to adapt in real time to objects referenced through natural language, regardless of whether they were part of the original environment model.

The resulting QP is solved at each control loop to produce the desired joint velocity vector \( \dot{\mathbf{q}}_d \), which satisfies the stacked tasks and all active inequality constraints. We employ an open-loop control scheme, initialized with the robot's generalized coordinates $ \mathbf{q} = [p_x, p_y, \psi, \bm{\theta}^\top]^\top \in \mathbb{R}^{3+n}$, where \( p_x \) and \( p_y \) denote the robot base position in the plane, \( \psi \) is the yaw orientation around the vertical axis, and \( \bm{\theta} \in \mathbb{R}^n \) represents the joint positions of the robot. At each control step, the QP output \( \dot{\mathbf{q}}_d \in \mathbb{R}^{3+n} \) is integrated to update the internal model state.

Cartesian commands from the VR interface are specified as homogeneous transformation matrices $\mathbf{T}_r \in \mathbb{R}^{4 \times 4}$. Orientation errors are computed using quaternion difference and scaled by a gain matrix $\mathbf{K}_C \in \mathbb{R}^{6 \times 6}$ before being combined with feedforward velocity inputs $\mathbf{v}_r \in \mathbb{R}^3$ and $\mathbf{w}_r \in \mathbb{R}^3$ from the VR controller.

\section{EXPERIMENTAL DEMONSTRATION}

\begin{figure*}[bhtp!]
    \centering
    \includegraphics[width=0.8\textwidth]{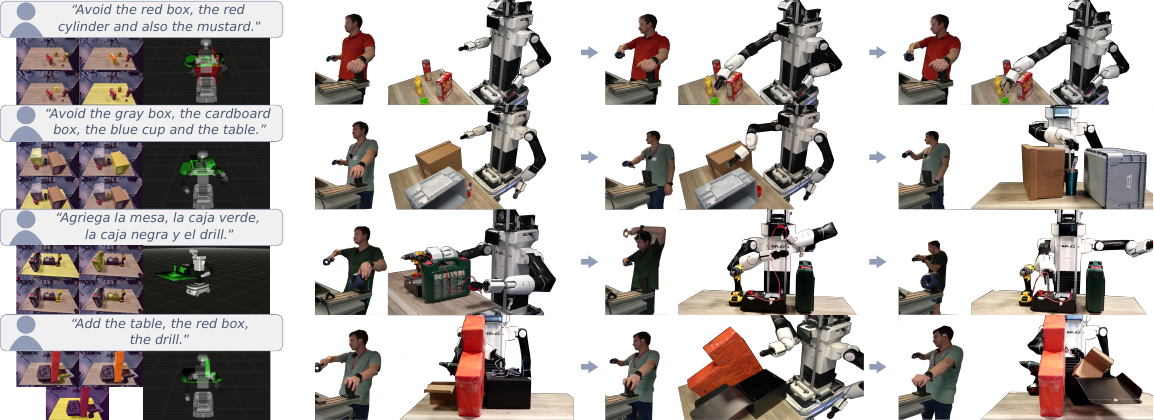}
    \vspace{-1em}
    \caption{The robot receives a spoken instruction from the operator. On the left, we show the 2D segmentation masks and the corresponding 3D collision avoidance meshes generated in response to the instruction. From left to right, the remaining images show snapshots of the teleoperator and the robot during task execution. A video demonstrating the experiments is available at: \url{https://hucebot.github.io/vlm_safe_teleop_website}}
    \label{fig:downstream_tasks}
\end{figure*}

We conducted a pilot study with five participants (volunteer adults, $1$ female and $4$ male, aged $22 \pm 1$ min 21, max 30) to quantitatively and qualitatively evaluate the performance of the proposed teleoperation system on pick-and-place tasks within narrow and cluttered environments. Prior to the experiment, all participants received standardized training to familiarize themselves with the system. During the trials, each participant issued collision avoidance commands in their language of preference (Spanish, English, French, Italian and Arabic, all supported by the multilingual whisper model) to test the system's robustness to linguistic variability.

The results demonstrate that the system successfully identified all objects for which collision avoidance was to be activated, irrespective of the language used and the way the participants chose to refer to the objects. Throughout the live trials, no collisions were recorded, confirming the system's effectiveness in collision prevention.

To assess the impact of the collision avoidance module, the recorded control commands from the handheld devices were subsequently replayed on the robot with collision avoidance disabled. In this condition, collisions occurred in four out of five trials, highlighting the significant contribution of the collision avoidance mechanism to task safety.

Representative examples of complex downstream tasks, performed outside of the pilot study, are illustrated in Fig.~\ref{fig:downstream_tasks} and also demonstrated in the supplementary video. The pilot study results, including detailed teleoperation sessions and participant interactions, can be seen in the video: \url{https://hucebot.github.io/vlm_safe_teleop_website}. This separation highlights both the controlled evaluation through the pilot study and the broader applicability of the system demonstrated in diverse manipulation scenarios.

\section{DISCUSSION AND FUTURE WORK}

Our system demonstrates how vision-language grounding can enable safe and intuitive bimanual teleoperation with collision avoidance. Its modular design ensures broad applicability, allowing rapid deployment across different robot platforms with minimal configuration.

Despite these strengths, several limitations suggest clear directions for future improvement.

A primary challenge lies in the reliance on a single, head-mounted RGB-D camera for perception. This placement is prone to occlusions during tabletop manipulation and often yields incomplete point clouds, particularly in dynamic environments. Incorporating additional sensors, such as wrist-mounted cameras, would provide multi-view observations and improve scene reconstruction. Fusing data from multiple perspectives could significantly enhance the completeness of object segmentation and tracking.

Another limitation stems from the current interaction paradigm, which requires users to explicitly specify each object to avoid via verbal commands. While this offers fine-grained control, future work could explore learning-based approaches to anticipate obstacles based on prior experience. Instead of requiring explicit enumeration, the system could infer relevant obstacles dynamically from high-level task descriptions, leveraging VLMs to reason about context and intent. This would enable a more abstract and flexible interaction style, where the operator specifies task goals and the system autonomously determines which elements to avoid.

Additionally, the current requirement to press a button to activate the microphone interrupts the teleoperation flow. Replacing this with a voice-activated trigger would allow operators to issue commands more naturally and seamlessly, further enhancing the fluidity of interaction.

Finally, although our framework supports control of the robot's omnidirectional base, our current experiments focused solely on upper-body manipulation. Future work will investigate tasks that combine base navigation and manipulation, highlighting the full capabilities of the system in more dynamic and mobile scenarios.

\section*{ACKNOWLEDGMENTS}
This work was supported by the EU Horizon project euROBIN (GA n.101070596), the France 2030 program through the PEPR O2R projects AS3 and PI3 (ANR-22-EXOD-007, ANR-22-EXOD-004), the Agence Innovation Défense (ATOR project), the CPER CyberEntreprises, and the Creativ'Lab platform of Inria/LORIA. It also benefited from government funding managed by the French National Research Agency under France 2030 via the ENACT AI Cluster (ANR-23-IACL-0004).


\bibliographystyle{IEEEtran}
\bibliography{ref}

\end{document}